\documentclass{article}
\usepackage{spconf,amsmath,graphicx}

\usepackage{amssymb}
\usepackage{booktabs}

\usepackage{multicol,multirow}
\usepackage{pifont}
    \newcommand{\cmark}{\ding{51}}%

\usepackage[table]{xcolor}
\usepackage{color, colortbl}
\usepackage{wrapfig}
\definecolor{Gray}{gray}{0.8}

\usepackage{caption} 
\usepackage{subcaption} 
 
\usepackage[hidelinks,pagebackref]{hyperref}
\hypersetup{
  colorlinks   = true, 
  urlcolor     = blue, 
  linkcolor    = blue, 
  citecolor   = red 
}

\title{Stability Plasticity Decoupled Fine-tuning For Few-shot end-to-end Object Detection}
\name{Yuantao Yin, Ping Yin*\thanks{* means the corresponding author.}}
\address{Inspur \\
generalvision@foxmail.com, generalvisionyyt@gmail.com}

\begin{document}
%
\maketitle
\begin{abstract}
Few-shot object detection(FSOD) aims to design methods to adapt object detectors efficiently with only few annotated samples. 
Fine-tuning has been shown to be an effective and practical approach. However, previous works often take the classical base-novel two stage fine-tuning procedure but ignore the implicit stability-plasticity contradiction among different modules. Specifically, the random re-initialized classifiers need more plasticity to adapt to novel samples. The other modules inheriting pre-trained weights demand more stability to reserve their class-agnostic knowledge. Regular fine-tuning which couples the optimization of these two parts hurts the model generalization in FSOD scenarios. In this paper, we find that this problem is prominent in the end-to-end object detector Sparse R-CNN for its multi-classifier cascaded architecture. We propose to mitigate this contradiction by a new three-stage fine-tuning procedure by introducing an addtional plasticity classifier fine-tuning(PCF) stage. We further design the multi-source ensemble(ME) technique to enhance the generalization of the model in the final fine-tuning stage. Extensive experiments verify that our method is effective in regularizing Sparse R-CNN, outperforming previous methods in the FSOD benchmark.
\end{abstract}
\section{Introduction}
\label{sec:intro}
As a fundamental task in computer vision, object detection is applied in various scenarios. Existing object detection methods\cite{sun2021sparse,ren2015faster} rely on sufficient annotated samples to perform well, which may not be available in practical applications. They still suffer from poor generalization or over-fitting when only few annotated examples are offered.

Few-shot object detection(FSOD) is proposed to focus on how to efficiently adapt to novel downstream tasks with limited instances. Prior work TFA\cite{Wang2020} propose that fine-tuning is a simple yet effective but ignored approach to adapt detectors if freezing protection is conducted to prevent over-fitting. In TFA, a two-stage transfer procedure is proposed and is popularly used by many FSOD methods. In the first stage, sufficient annotated instances which belong to base classes are available for pre-training the detector. In the following stage, only few fixed number of annotated instances of each novel class are available for adapting the detector. Base classes and novel classes have no intersection.

\begin{figure}[t]
\centering
\includegraphics[width=8.0cm]{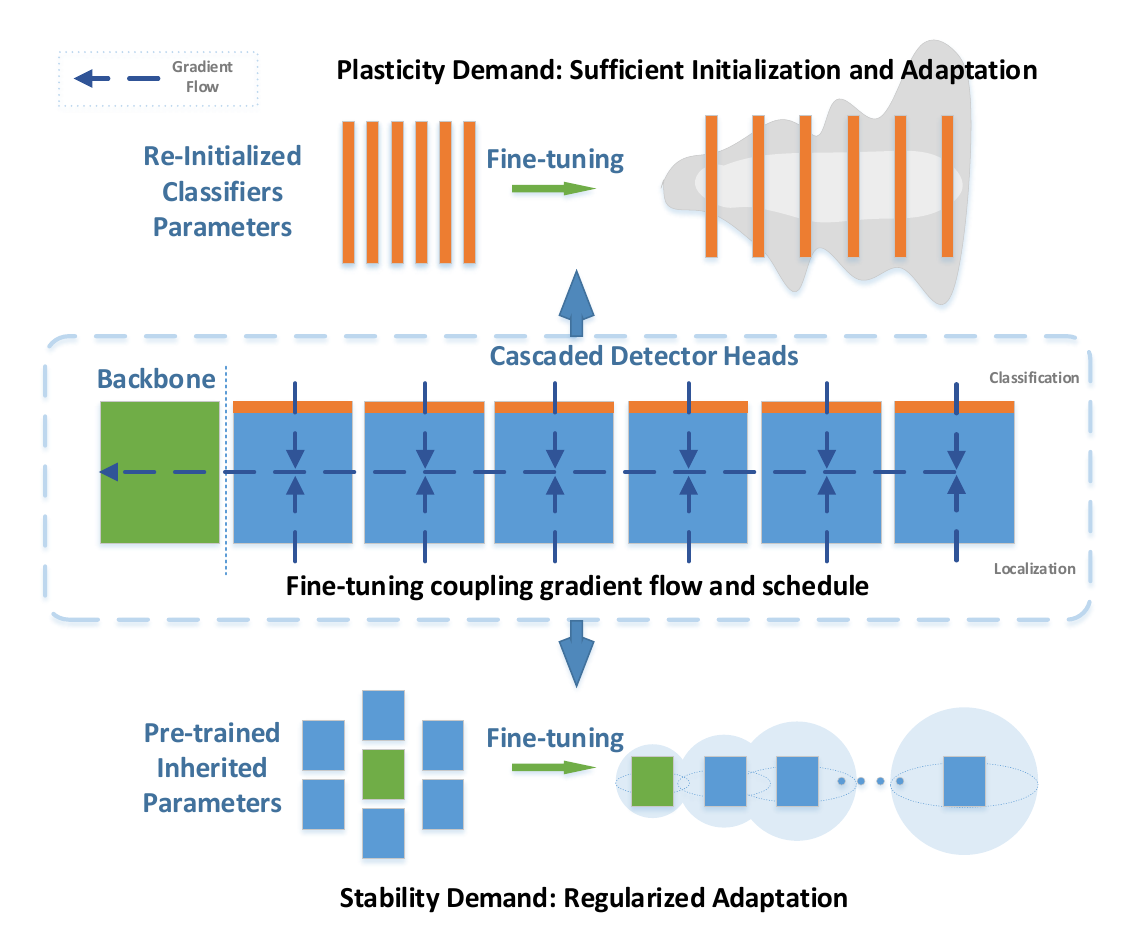}
\caption{Stability-Plasticity contradiction in Sparse R-CNN\cite{sun2021sparse}. During fine-tuning in regular two-stage procedure, re-initialized classifiers and other modules actually have contradictory stability-plasticity demand. However, fine-tuning coupling them together can only achieve sub-optimal FSOD performance for Sparse R-CNN.}
\label{fig:res}
\end{figure}

Many methods use this two-stage transfer procedure and take Faster R-CNN\cite{ren2015faster} as their detector whose architecture is effective and interpretable. They find and tackle many problems during fine-tuning from various views. MPSR\cite{wu2020multi} tackles the scale variation problem. FSCE\cite{Sun2021} updates the freezing strategy and propose a contrastive loss to alleviate confusion. FADI\cite{Cao2021} presents an association and discrimination loss based on the semantic similarity. However, the limitations of this two-stage transfer procedure is still less discussed. And there is no work explore the fine-tuning efficiency from the perspective of implicit plasticity-stability contradiction among different modules.


Some end-to-end detectors\cite{carion2020end,sun2021sparse} are proposed and perform competitively recently. Among them, typically, the design of Sparse R-CNN\cite{sun2021sparse} has the advantages of efficient training convergence and high interpretability. It takes an relatively explicit object encoding correlation method by interacting with backbone feature maps through ROI pooling\cite{ren2015faster}. These attract us to explore the FSOD performance of it. Similar with DETR\cite{carion2020end}, the architecture of Sparse R-CNN consists of a stack of cascaded detector heads which refine the results progressively. Each head has a pair of linear classifier and regressor to predict the category and localization of instances. Thus the parameter number of these multiple linear classifiers are much more than Faster R-CNN which has only single linear classifier, considering the number of cascade stages is often 6 in Sparse R-CNN. 

In this paper, we find the stability-plasticity contradiction between these linear classifiers and other modules in FSOD, inspired by stability-plasticity dilemma in continual learning works\cite{wang2023comprehensive}. The FSOD performance of Sparse R-CNN suffer much from this contradiction due to its multi-classifier architecture. As shown in Fig.\ref{fig:res}, when we fine-tune Sparse R-CNN with the two-stage procedure, after the first base stage, we have to re-initialize the weights of all these linear classifiers to random values due to the change of class number. Thus these classifiers demand more plasticity to initialize first and then adapt to novel tasks during optimization. This demand is greater when multiple classifiers exist. In contrast, other modules just optimize to adapt the pre-trained weights with novel samples. Considering the much larger number of parameters, they require more stability to avoid the risk of over-fitting. However, these two parts of modules with inconsistent stability-plasticity demand are unreasonably coupled during optimization in the novel fine-tuning stage of the popularly used transfer procedure by previous works. This compromising optimization often converges to the model weights whose output features are distorted\cite{kumar2021fine}. Among them, classifiers tend to be under-fitted while others tend to be over-fitted simultaneously. Thus this coupling scheme hurts the generalization of the detector in FSOD. And this problem is especially prominent in Sparse R-CNN for its much larger number of parameters in classifiers.

To address the above issues, we propose to split the classical two-stage transfer procedure into a new three-stage one. Speicifically, following the base stage, an exclusive stage which focuses on fine-tuning classifiers in Sparse R-CNN sufficiently is introduced. In this stage, we freeze all modules except these re-initialized classifiers of these cascaded heads. And we set a sufficient long schedule to initialize and adapt these classifiers with novel samples. This meets the demand of plasticity of classifiers without concerning the risk of over-fitting. Then in the following novel fine-tuning stage, we propose an ensemble aggregation method to reserve the generalization stability of the detector. We regularise the detector by aggregating prediction results predicted by models with weights which are trained in diverse configurations. Extensive experiments are conducted to verify the validity of these measures to boost FSOD performance of Sparse R-CNN against its baseline.



\section{Related Works}
\label{sec:related}
\subsection{Few-Shot Object Detection}
There are two branches in the FSOD research field. One is the meta-learning branch. The typical methods includes FSRW\cite{Kang2019}, Meta-RCNN\cite{Yan2019}, FsDetView\cite{xiao2020few}, Meta-DETR\cite{zhang2021meta}. The other is the fine-tuning transfering branch. The typical works are TFA\cite{Wang2020}, DeFRCN\cite{Qiao2021}, FADI\cite{Cao2021}. DeFRCN\cite{Qiao2021} discovers the implicit multi-stage and multi-task contradiction which is exaggerated in the FSOD setting and propose affine adapter layers. Recently, some FSOD works start to utilize end-to-end detectors. Meta-DETR\cite{zhang2021meta} incorporates correlational aggregation for meta-learning with the Deformable DETR\cite{zhu2020deformable} detector. COCO-RCNN\cite{ma2022few} proposes contrastive loss to regularise Sparse R-CNN to improve its consistent concentration.

Our approach follows the paradigm of fine-tuning, and take an end-to-end paradigm model Sparse R-CNN as our detector. But we focus on analysing its implicit plasticity-stability contradiction and propose corresponding techniques to alleviate it, which is still not discussed in previous FSOD works. 

\subsection{Stability-plasticity Dilemma}
The stability-plasticity dilemma is a well-known contradiction in both artificial and biological neural systems. Unbalanced stability-plasticity strategy in continual learning settings often lead to catastrophic forgetting when the model learn multiple tasks in sequence. If memory stability is emphasized too much, the model suffers from low efficiency to adapt to novel tasks. Vice versa, if learning plasticity has no regularization constraints, the neural network disrupts the previous knowledge when assimilate new information\cite{parisi2019continual}. In continual learning, Some works\cite{kirkpatrick2017overcoming,aljundi2018memory} solves this dilemma by weight regularization. Some works\cite{gurbuz2022nispa,wang2022coscl,serra2018overcoming} explores the modular-based approach. 

In this work, as far as we know, it is the first time to introduce this stability-plasticity perspective into FSOD setting. Here we focus on not protecting knowledge of previous tasks but on reserving class-agnostic generalization knowledge from upstream pre-training. To differentiate with the concept in continual learning, we call it as stability-plasticity contradiction here.


\section{Methods}
\label{sec:methods}
\subsection{Preliminaries}
We clarify the two-stage transfer procedure proposed by the previous work\cite{Wang2020} first and then formalize the pipeline of Sparse R-CNN.

In the base stage, a base dataset $D_{base}$ with abundant annotated instances whose classes is $C_{base}$ is offered for transferable knowledge extraction. In the novel stage, a novel dataset $D_{novel}$ with limited but balanced K-shot annotated instances is offered for fine-tuning transfer. The classes of the instances in $D_{novel}$ are $C_{novel}$. And the base classes $C_{base}$ in $D_{base}$ are strictly exclusive with the novel classes $C_{novel}$ in $D_{novel}$, namely, $C_{novel} \cap C_{base} = \phi$. When evaluating, a test dataset $D_{test}$ is offered. The classes of the annotated instances in $D_{test}$ is $C_{novel}$.  

The detection pipeline of Sparse R-CNN can be formulated as:
\begin{equation}\label{sprequ}
\begin{array}{llr}
    x^{FPN} &= B_{FPN}(I), &(1a) \\
    x_{i} &= R_{pool}(x^{FPN}, b_{i-1}), &(1b) \\
    q_{i} &= A_{i}(q_{i-1}, x_{i}),&(1c)\\ 
    u_{i} &= D_{i}^{box}(q_{i}),&(1d)\\
    v_{i} &= D_{i}^{cls}(q_{i}),&(1e)\\
    b_{i} &= F_{i}^{box}(u_{i}),  &(1f)\\
    c_{i} &= F_{i}^{cls}(v_{i}), & (1g)
\end{array}
\end{equation}
where $q$ denotes object encoding vectors. $x^{FPN}$ denotes the output features of $B_{FPN}$ the backbone model with FPN neck. $b$ denotes the predicted box. $c$ denotes the predicted classification label. $R_{pool}$  denotes RoI pooling operation. $A$ denotes the attention correlation module. $D$ 
denotes the modules decoding object encodings into regression or classification features. $F^{cls}$ denotes the linear classifier. $F^{box}$ denotes the linear regressor. $i\in\{1...N_H\}$. $N_H$ is often 6. $b_0$ and $q_0$ are trainable parameters in the model.

\subsection{Plasticity Classifier Fine-tuning}
We propose to insert an independent plasticity classifier fine-tuning(PCF) initialization stage after the base stage and before the novel stage for Sparse R-CNN during fine-tuning in the FSOD setting. In this stage, the available samples are the same with the novel stage. The regular base-novel procedure is updated to a base-init-novel procedure.

We define $\mathcal{C}=\{\mathcal{P}(F_{i}^{cls})|i\in{1...N_H}\}$ as the parameters of all linear classifiers in all stages in Sparse R-CNN. $\mathcal{P}(\mathbf{\cdot})$ denotes the operator to get the parameters of the input module. We define $\mathcal{E}=\{\mathcal{P}(B_{FPN},A_i,D_i^{box},D_i^{cls},F_i^{box})|i\in{1...N_H}\}$ as the parameters of all other modules in Sparse R-CNN.

At the beginning of this stage, $\mathcal{C}$ are all random initialized. $\mathcal{C}$ has no demand to reserve inherited knowledge. It requires sufficient initialization and adaptation during transfer instead. Therefore, sufficient plasticity is necessary to be offered to facilitate its optimization convergence. Otherwise, $\mathcal{C}$ under-fits to $D_{novel}$.

Since the number of available samples $D_{novel}$ is rather limited, these samples have strong sample selection bias. Considering $C_{novel} \cap C_{base} = \phi$, so they have both out-of-distribution shifts from $D_{test}$ and from $D_{base}$. Therefore, the gradients generated with $D_{novel}$ during fine-tuning can only adapt features which are not orthogonal to the distribution of $D_{novel}$. Positive transferable knowledge extracted from distribution of $D_{base}$ in $\mathcal{E}$ which can generalize to $D_{test}$ but are orthogonal to the subspace of $D_{novel}$ is easy to be distorted and forgotten\cite{kumar2021fine}. $\mathcal{E}$ over-fits to $D_{novel}$ if no protection constraints are imposed.

Moreover, Sparse R-CNN takes the cascaded architecture rather than the parallel one like Faster R-CNN. The gradient flow through $\mathcal{C}$ and $\mathcal{E}$ are coupled during optimization. Moreover, the gradient descent optimization tend to automatically balance the weights between different layers\cite{du2018algorithmic}. So during the plasticity classifier optimization, parameters in $\mathcal{E}$ deviates from its pre-trained weights concomitantly with the optimization of $\mathcal{C}$ if $\mathcal{E}$ is not constrained. This leads to severe forgetting of positive transferable knowledge in $\mathcal{E}$. 

Considering these, in this exclusive stage, we decouple the optimization schedule to offer high plasticity to $\mathcal{C}$ but protect $\mathcal{E}$ strictly. We separate the necessary adaptation of $\mathcal{E}$ from here and leave it to the following stage. We set long optimization schedule to assure the sufficient fitting of $\mathcal{C}$ without worrying the over-fitting of $\mathcal{E}$. No regularization and constraints are imposed to $\mathcal{C}$ while full protection is conducted on parameters in $\mathcal{E}$ by freezing. The checkpointed model weights of this stage are then used in the following step.

\subsection{Stability Regularisation}
Class-specific and class-agnostic knowledge are mixed in parameters $\mathcal{E}$. Apparently full freezing $\mathcal{E}$ in the previous stage is too strong that necessary class-specific adaptation to novel tasks of $C_{novel}$ is inhibited. So following that, the novel fine-tuning is still necessary to adapt these class-specific knowledge further. On the other hand, considering the stability demand of $\mathcal{E}$ to reserve the class-agnostic generalization of the original pre-trained weights. During this fine-tuning, we freeze the two bottom blocks of backbone $B_{FPN}$. Furthermore, we propose a method called Multi-source Ensemble(ME) to regularise the detector after this adaptation.

Inspired by previous works\cite{Snell2017,wortsman2022robust,Qiao2021}, model weights before fine-tuning can often provide extra diversity and robustness to rectify the weights deviation and increase the generalization. WISE-FT\cite{wortsman2022robust} demonstrates effective generalization ensemble method from the weight space view by linearly interpolating weights before and after fine-tuning. However, this view relies on mode connectivity theory\cite{frankle2020linear} to require the model to have zero-shot classifiers, which is not satisfied here. 

Here we propose an ensemble regularization method to further regularize the transfer.
We introduce an array of prototypes using the offered ImageNet pre-trained backbone model $B^{Im}$ which is neither fine-tuned nor pre-trained in the base stage. The image $x_j^{sp}$ is input into $B^{Im}$ to extract the feature map. Then we crop the representation vector $u_j$ of each instance by using ROIAlign with $b_j^{sp}$ to get $u_j$. We calculate the prototypes $\mathcal{Q}=\{r_c^{Im}|c\in\{1...N\}\}$ as:
\begin{equation}
r_c^{Im}=\frac{1}{\left|\mathcal{S}_c\right|}\sum_{y_j^{sp}\in\mathcal{S}_c}u_j, \text{    where    } u_j \gets \mathcal{R}(B^{Im}(x_j^{sp}),b_j^{sp})
\end{equation}
where the operator $\mathcal{R}(\cdot)$ denotes ROIAlign.

Given an object proposal $\hat{y}_k=$ $(c_k,s_k,b_k)$ predicted by the detector with image $\hat{x}_k$ during inference, where $c_k,s_k,b_k$ denotes the category label,confidence score and bounding box correspondingly. We get its feature $z_k^{Im}=R_{pool}(B^{Im}(\hat{x}_k), b_k)$ using the member $B^{Im}$. Then we calculate its cosine similarity with prototypes in $\mathcal{Q}$ as: $s_k^{Im}=cos(z_k^{Im}, r_c^{Im})$, where the operator $cos(\cdot)$ denotes cosine similarity operation. At last, we ensemble $s_k^{Im}$ with $s_k$ to rectify the prediction results $s_k$ of the detector as:
\begin{equation}
\begin{array}{c}
    s^{ens}=\alpha\cdot s_k + \beta\cdot s_k^{Im} ,\\
    \alpha+\beta=1
\end{array}
\end{equation}
where $\alpha,\beta$ are hyper-parameters to aggregate the ensemble results from each member.


\section{Experiments}
\label{sec:experiments}
\begin{table}[t]
\centering
\resizebox{\linewidth}{!}{
    \begin{tabular}{l|c|cccccc}
    \toprule
    \multicolumn{1}{c|}{\multirow{2}[2]{*}{\shortstack{Method \vspace{2pt}}}} &\multicolumn{1}{c|}{\multirow{2}[2]{*}{\shortstack{BackboneDet \vspace{2pt}}}} &\multicolumn{6}{c}{Shot Number} \\
          & &\multicolumn{1}{c}{1} & \multicolumn{1}{c}{2} & \multicolumn{1}{c}{3} & \multicolumn{1}{c}{5} & \multicolumn{1}{c}{10} & \multicolumn{1}{c}{30} \\
    \midrule
    TFA w/cos\cite{Wang2020} &FRCN-R101 &3.4     & 4.6   & 6.6   & 8.3     & 10.0   & 13.7 \\
    MPSR\cite{wu2020multi}&FRCN-R101 & 2.3     & 3.5   & 5.2   & 6.7     & 9.8   & 14.1 \\
    Attention-RPN\cite{fan2020few}&FRCN-R101 & 4.2     & 6.6   & 8.0   & 6.7     & 9.8   & 14.1 \\
    FSCE\cite{Sun2021}&FRCN-R101 & -     & -   & -   & -     & 11.1   & 15.3 \\
    FADI\cite{Cao2021} & FRCN-R101 & 5.7 & 7.0 & 8.6 & 10.1 & 12.2 & 16.1 \\
    TIP\cite{li2021transformation} & FRCN-R101 & 5.7 & 7.0 & 8.6 & 10.1 & 12.2 & 16.1 \\
    Meta Faster R-CNN\cite{han2022meta} & FRCN-R101 & 5.1 & 7.6 & 9.8 & 10.8 & 12.7 & 16.6 \\
    \midrule
    SPRCNN-ft-full &SPR-R101 & 1.3   & 2.9  & 4.5  & 7.3  & 9.7 & 16.8 \\
    COCO-RCNN\cite{ma2022few}&SPR-R101 & 5.2   & -  & -  & -  & 16.4 & 19.2 \\
    \rowcolor{Gray}
    \textbf{Ours} &SPR-R101 & 6.5   & 9.6  & 11.7  & 14.7  & 17.6 & 23.1 \\
    \bottomrule
    \end{tabular}}%



\caption{\label{benchmarkres}Performance comparison with the baseline and some previous FSOD methods on COCO benchmark(novel $mAP$). We list their backbone models and detectors and align them in separate groups for fair comparison.}
\end{table}

\subsection{Datasets}
We use the dataset MS COCO-2014 as in works \cite{Cao2021,Wang2020,Sun2021} to evaluate the FSOD performance of our method. For COCO dataset, the 60 categories non-overlapping with PASCAL VOC dataset are treated as base classes, and the remaining 20 classes are selected as novel classes. We report the detection accuracy for AP,AP50,AP75 on the test set consisting of 5k images with shot settings at 1,2,3,5,10 and 30. 

\begin{table}[t]
\centering
\resizebox{0.7\linewidth}{!}{
\begin{tabular}{c|cc|ccc}
\toprule
Detector &  \multicolumn{2}{c}{Method}  & \multicolumn{3}{c}{10shot}   \\
\midrule
SPRCNN &  PCF     & ME  & AP & AP50 & AP75 \\
\midrule
\cmark     &      &    &      9.7 & 15.9 & 9.7 \\
\cmark     & \cmark      &   &    13.5 & 21.0 & 13.9 \\
\cmark     &  & \cmark   &    11.9 & 19.3 & 11.9 \\
\cmark     & \cmark     &   \cmark &    17.7 & 27.8 & 18.0 \\
\bottomrule
\end{tabular}
}
\medskip
\caption{\label{ablationtab}Ablation study of the effectiveness of proposed methods in 10-shot setting on MS-COCO benchmark. PCF, ME denotes Plasticity Classifier Fine-tuning, and Multi-source Ensemble respectively.}
\end{table}

\subsection{Implementation Details}
We use the typical end-to-end object detecotr Sparse R-CNN\cite{sun2021sparse} as the base framework. And we use ResNet-101\cite{he2016deep} as our backbone model as previous works. AdamW is utilized as the optimizer with the weight decay of $e^{-4}$. We use a batch size of 8 on four GPUs. 

\subsection{Benchmark Results}
We compare the proposed method with previous works and show the results in Tab.\ref{benchmarkres}. For fairness consideration, we show the baseline performance by fully fine-tuning Sparse R-CNN without any regularization to rule out the difference brought by the detector itself. Our method achieves improvements $7.2\%,7.4\%,7.9\%,6.3\%$ mAP in 3,5,10,30-shot settings respectively in comparison with its baseline performance. 


\subsection{Ablation Study}
Ablation experiments are conducted on 10 shot settings of the MS-COCO benchmark to analyze the contributions of each component of the proposed methods. 

All results are shown in Tab.\ref{ablationtab}. Specifically, we see in the first row of this table that if we fine-tune plain Sparse R-CNN naively without any proposed techniques, it only achieves $9.7\%$ mAP for the 10-shot setting. This poor FSOD performance indicates that the existence of implicit plasticity-stability contradiction in the detector Sparse R-CNN leads to severe over-fitting to the limited samples. 

Follwing that, we see if we introduce the plasticity classifier fine-tuning technique, the FSOD performance improves dramatically from $9.7\%$ to $13.5\%$  mAP on this 10-shot setting. This shows decoupling the plasticity fine-tuning stage from the regular novel fine-tuning stage is greatly beneficial in balancing the stability-plasticity trade-off to reserve more generalization robustness of the detector. 

Next, in the third and fourth line, we can see the effectiveness of the proposed multi-source ensemble method independently and when it works jointly with the plasticity classifier fine-tuning method. It  pushes the FSOD performance from $9.7\%$ to $11.9\%$ independently by mitigating the negative deviation of the detector from its pre-trained weights. When applied jointly, 10-shot FSOD performance is further improved to $17.7\%$ mAP.

\section{CONCLUSION}
\label{sec:conclusion}
In this paper, we analyze the stability-plasticity contradiction between linear classifiers and other modules in Sparse R-CNN during Fine-tuning in the FSOD scenario. We propose a new three-stage transfer procedure by introducing an additional classifier plasticity fine-tuning stage for Spase R-CNN. Moreover, we design a an ensemble technique to further regularise the detector. The implicit stability plasticity contradiction is mitigated with our method. Experiments verify the effectiveness of the proposed method.


\bibliographystyle{IEEEbib}
\bibliography{strings,refs}

\end{document}